\documentclass[journal,twoside,web]{IEEEtran}
\usepackage{cite}
\usepackage{amsmath,amssymb,amsfonts}
\usepackage{algorithmic}
\usepackage{graphicx}
\usepackage{algorithm,algorithmic}
\usepackage{hyperref}
\hypersetup{hidelinks=true}
\usepackage{textcomp}

\usepackage{color}

\def\BibTeX{{\rm B\kern-.05em{\sc i\kern-.025em b}\kern-.08em
    T\kern-.1667em\lower.7ex\hbox{E}\kern-.125emX}}
\begin{document}

\title{Improving  Risk Stratification in Hypertrophic Cardiomyopathy:  A Novel Score Combining Echocardiography, Clinical, and Medication Data}

\author{Marion Taconné$^1$, 
        Valentina D.A. Corino$^{1,2}$, Annamaria Del Franco$^3$, Sara Giovani$^3$,  Iacopo Olivotto$^3$, Adrien Al Wazzan$^4$, Erwan Donal$^4$, 
        Pietro Cerveri$^{1,5}$,  
		Luca Mainardi$^1$,~\IEEEmembership{Member,~IEEE}
\thanks{This work was supported by the European Union through the project SMASH-HCM under grant 101137115 and INSIGHT-LV under grant 101198472 }
\thanks{$^{1}$Marion Taconné, and Luca Mainardi are with the Department of Electronics, Information and Bioengineering (DEIB), Politecnico di Milano, 20133 Milano, Italy. {\tt\small marionhelene.taconne@polimi.it}; 
{\tt\small luca.mainardi@polimi.it}.}
\thanks{$^{1,2}$Valentina D.A. Corino is with the Department of Electronics, Information and Bioengineering (DEIB), Politecnico di Milano, 20133 Milano, Italy, and also with the CardioTech Lab, IRCCS Centro Cardiologico Monzino, 20138 Milano, Italy {\tt\small  valentina.corino@polimi.it}.}%
\thanks{$^{3}$Annamaria Del Franco, Sara Giovani and Iacopo Olivotto are with Cardiomyopathy Unit, Careggi University Hospital, 50134 Florence, Italy.
{\tt\small annamaria.delfranco@gmail.com}; {\tt\small giovani.sara@yahoo.com}; {\tt\small iacopo.olivotto@unifi.it}.}
\thanks{$^{4}$Adrien Al Wazzan, and Erwan Donal are with Univ Rennes, CHU Rennes, Inserm, LTSI – UMR 1099, 35000 Rennes,  France.
{\tt\small adrien.al.wazzan@chu-rennes.fr}; {\tt\small erwan.donal@chu-rennes.fr}.}
\thanks{$^{1,5}$ Pietro Cerveri is with the Department of Electronics, Information and Bioengineering (DEIB), Politecnico di Milano, 20133 Milano, Italy, and also with the Department of Electrical, Computer and Biomedical Engineering, University of Pavia, 27100 Pavia, Italy {\tt\small pietro.cerveri@unipv.it}.}
}

\maketitle

\thispagestyle{empty} 
\pagestyle{plain}

\begin{abstract}
Hypertrophic cardiomyopathy (HCM) requires accurate risk stratification to inform decisions regarding ICD therapy and follow-up management. Current established models, such as the European Society of Cardiology (ESC) score, exhibit moderate discriminative performance. 

This study develops a robust, explainable machine learning (ML) risk score leveraging routinely collected echocardiographic, clinical, and medication data, typically contained within Electronic Health Records (EHRs), to predict a 5-year composite cardiovascular outcome in HCM patients. The model was trained and internally validated using a large cohort ($N=1,201$) from the SHARE registry (Florence Hospital) and externally validated on an independent cohort ($N=382$) from Rennes Hospital. 

The final Random Forest ensemble model achieved a high internal Area Under the Curve (AUC) of $0.85 \pm 0.02$, significantly outperforming the ESC score ($0.56 \pm 0.03$). Critically, survival curve analysis on the external validation set showed superior risk separation for the ML score (Log-rank $\text{p} = 8.62 \times 10^{-4}$) compared to the ESC score ($\text{p} = 0.0559$). Furthermore, longitudinal analyses demonstrate that the proposed risk score remains stable over time in event-free patients. 

The model’s high interpretability and its capacity for longitudinal risk monitoring represent promising tools for the personalized clinical management of HCM.
\end{abstract}

\begin{IEEEkeywords}
Explainable AI (XAI), Hypertrophic cardiomyopathy (HCM), Longitudinal analysis, Machine Learning (ML), Risk stratification, Electronic Health Records (EHR), Echocardiography

\end{IEEEkeywords}

\section*{Disclaimer}

This work has been submitted to the IEEE for possible publication. Copyright may be transferred without notice, after which this version may no longer be accessible.

\section{Introduction}

\label{sec:introduction}

\IEEEPARstart{H}{ypertrophic} cardiomyopathy (HCM) is a complex genetic heart disease characterized by unexplained left ventricular (LV) hypertrophy, myocardial disarray, and interstitial fibrosis. It remains a major cause of sudden cardiac death (SCD), particularly among young individuals, with an estimated annual risk of approximately 2$1\%$ in unselected populations \cite{OMahony2014, Ommen2020}. In HCM patients, arrhythmic events arise from hypertrophied myocardial regions, where fibrosis and disarray create an arrhythmogenic substrate. Although exertion has traditionally been viewed as a common trigger, most sudden deaths occur during rest or mild activity, as shown in unselected cohorts \cite{Maron2000, Ashkir2025}.

Implantable cardioverter-defibrillators (ICDs) are effective in the primary prevention of SCD but involve non-negligible risks, including device-related complications, inappropriate shocks, and psychological distress \cite{Priori2015, Zeppenfeld2022, Arbelo2023}. Therefore, accurate identification of high-risk patients is critical to inform ICD implantation decisions and to optimize the balance between clinical benefit and potential harm.

Current international guidelines rely on established clinical risk models such as the 5-year SCD risk calculator developed by the European Society of Cardiology (ESC) \cite{OMahony2014, Zeppenfeld2022} and the risk factors defined by the American Heart Association/American College of Cardiology (AHA/ACC) \cite{Ommen2020}. These models are built on a limited set of predefined clinical variables and have demonstrated only moderate discriminative performance, with the ESC score reporting C-index values around $0.69$ in validation cohorts \cite{OMahony2014}.

Recent research in HCM has moved toward machine-learning (ML) approaches to improve these outcomes. Prior ML studies based on electrocardiography (ECG) have either focused on identifying distinct electrical phenotypes without direct outcome prediction \cite{Lyon2018}, or on linking ECG markers to underlying myocardial disarray and fibrosis to provide mechanistic insight \cite{Ashkir2025}, rather than on deployable longitudinal risk stratification. Zhao et al. \cite{Zhao2024} highlighted the prognostic value of combining echocardiography with cardiac magnetic resonance (CMR), while Lai et al. \cite{Lai2025} utilized multimodal EHR data to refine risk stratification. However, such approaches were either basically static, providing only a baseline risk, or relied on imaging modalities like CMR that are not universally accessible in routine workflows~(Table \ref{tab:prior_ml_approaches}).

Despite the central role of echocardiography in HCM diagnosis and management \cite{Franco2023, Maron2022, Mitchell2023, Maurizi2025}, its rich set of quantitative parameters is often underexploited in existing prognostic tools. Key features like LV diameter, maximum wall thickness, left ventricular outflow tract (LVOT) gradient, and left ventricular ejection fraction (LVEF) are integral to current scores. However, parameters derived from the ECG and medication use are typically not included in these established risk models, despite the potential of ECG markers to reflect arrhythmogenic risk and medication status to reflect disease management.

\begin{table*}[t]
\caption{Summary of prior machine-learning approaches for risk stratification in hypertrophic cardiomyopathy (HCM), highlighting quantified measures, model outputs, and clinical limitations.}
\label{tab:ml_hcm_related_work}
\centering
\renewcommand{\arraystretch}{1.2}
\begin{tabular}{p{1cm} p{4cm}  p{5cm} p{5cm}}
\hline
\textbf{Study} & \textbf{Measures} & \textbf{Output} & \textbf{Clinical flaws} \\
\hline
\cite{Lyon2018}  
& ECG-derived electrical phenotypes 
& ECG pattern subgroups in HCM (unsupervised phenotyping)
& No direct risk stratification output; no time-to-event modeling; no longitudinal assessment of ECG evolution \\
\hline
\cite{Ashkir2025}  
& ECG-derived arrhythmic risk markers 
& Associations between ECG repolarization markers and myocardial disarray/fibrosis
& Mechanistic insight rather than patient-level risk; no risk stratification; No ICD decision support \\
\hline
\cite{Kochav2021} 
& Conventional echocardiography measurements; basic clinical variables
& Predicted probability or class of composite adverse cardiac events
& Static baseline risk estimation; no longitudinal risk evolution; limited generalizability \\
\hline
\cite{Zhao2024} 
& LGE extent and location; CMR feature-tracking strain; clinical and echocardiographic variables
& Static risk score for long-term major adverse cardiovascular events; stratification into low/medium/high risk groups
& Dependence on CMR imaging; static risk estimation without temporal updating \\

\hline
\cite{Lai2025} 
&  Myocardial fibrosis and scar patterns from LGE-CMR images; CMR-derived structural measures; Echo-derived features 
& Static probability of arrhythmic sudden cardiac death
& Computationally intensive; reliance on advanced CMR imaging; static risk estimation \\

\hline
\end{tabular}
\label{tab:prior_ml_approaches}
\end{table*}

In previous work \cite{Taconne2025a}, we developed a machine learning (ML)-based risk score that uses conventional echocardiographic features to predict a 5-year composite cardiovascular outcome in HCM patients. 
Building on these results, we address a critical gap: the lack of external validation and the limited exploration of the temporal evolution of ML risk.

In this work, we propose: i) an explainable ML risk score for hypertrophic cardiomyopathy integrating routinely available echocardiographic, clinical, and medication data typically contained within Electronic Health Records (EHRs); ii) a framework for model interpretability through Shapley additive explanations (SHAP) analysis~\cite{Salih2024}, identifying both established guideline-based predictors and additional echocardiographic markers; and iii) a longitudinal risk-monitoring framework that reconstructs patient-specific risk trajectories over repeated examinations. To our knowledge, this is the first study to combine routine clinical and medication data from EHRs with external validation and longitudinal slope analysis, providing a foundation for dynamic clinical management in HCM.

\begin{figure*}[t]
    \begin{center}
\includegraphics[width=18cm]{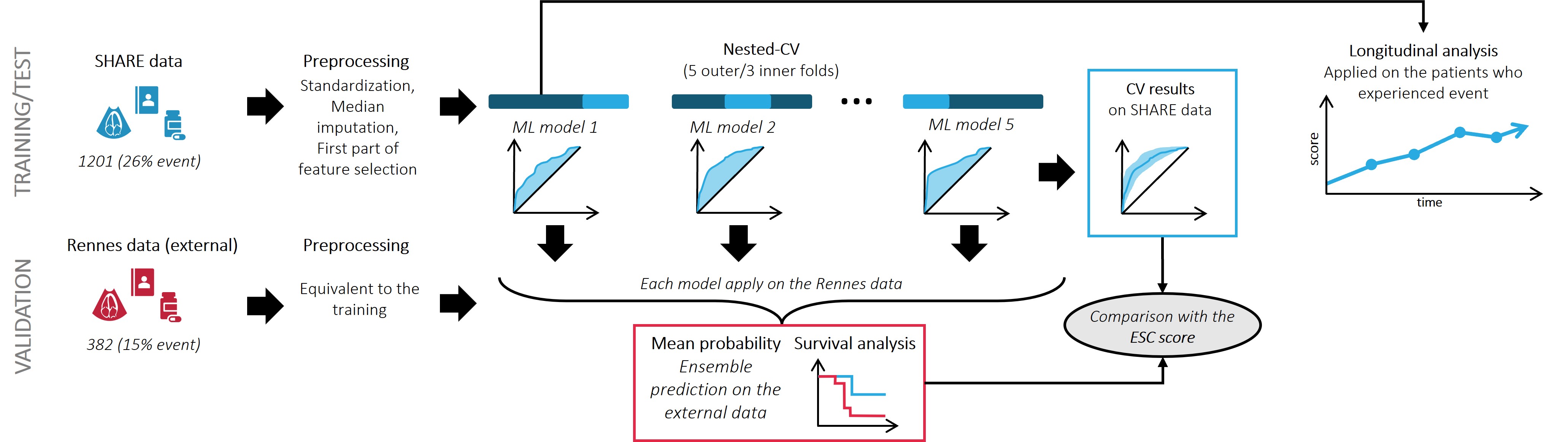}
    \end{center}
\caption{Methodological steps separated in training-test (top) and validation phase (bottom) with the survival analysis and longitudinal analysis. The five training models are gathered to form a final ensemble model.}
\label{fig:methodo_long_analysis}
\end{figure*}

\section{Methodology}

\subsection{Study population}

The initial study population consisted of 2,244 patients diagnosed with HCM at Careggi University Hospital (Florence, IT). To ensure sufficient data for 5-year risk prediction, we applied strict inclusion criteria: patients were required to have at least 5 years of follow-up or to have experienced a composite cardiovascular event within 5 years following a baseline echocardiographic examination. Additionally, patients without any recorded echocardiographic data were excluded. After the exclusion of cases with missing clinical data, the final primary dataset included 1,201 patients. For each patient, standard echocardiographic examinations and routine clinical assessments, including height, weight, and cuff blood pressure, were analyzed. Additional clinical information was collected on family history, medication use, and previous cardiovascular events, all of which were extracted from the EHRs.

An external validation dataset included 382 patients with HCM from Rennes University Hospital (Rennes, FR), were used. The mean follow-up period was $2.3 \pm 1.7$ years.

HCM was defined as unexplained left ventricular (LV) hypertrophy with a maximal LV wall thickness $>15$ mm, or $>13$ mm in first-degree relatives of HCM patients.

The composite outcome at 5 years comprises heart failure progression (including cardiac transplant, device implantation, LVEF $<$ 35\%, or worsening NYHA class to III-IV) and arrhythmic events (SCD, aborted cardiac arrest, appropriate ICD therapy). 
Patients were classified as high risk if at least one of these events occurred within five years after the baseline examination.
During the 5-year follow-up, 309 patients (25.7\%) in the Florence cohort and 57 patients (14.9\%) in the Rennes cohort experienced at least one event of the composite endpoint.

The predictor space initially comprised 82 distinct features. This included 63 echocardiographic measurements encompassing structural and functional dimensions, 14 clinical variables derived from physical examinations and patient history, and 5 features related to specific medication classes taken. Feature reduction primarily targeted the echocardiographic data to eliminate information redundancy and ensure compatibility with the external validation database, where certain features were unavailable.

\subsection{Feature selection and model training}

Features with pairwise correlation above 0.75 were removed, guided by clinical relevance. The remaining features were subjected to sequential floating forward selection (SFFS) \cite{Ferri1994}, implemented in Scikit-Learn \cite{scikit-learn}.
This SFFS was embedded in a nested cross-validation (CV) pipeline to ensure unbiased performance estimation and avoid data leakage. In the outer loop (5-fold stratified CV), each training set was used for model development, while an inner loop (3-fold stratified CV) performed hyperparameter optimization via grid search and applied SFFS.

Four classifiers were tested: random forest (RF), logistic regression (LR), support vector machine (SVM), and gradient boosting (GB), each with tailored hyperparameter grids. All data were preprocessed by median imputation, standardization, and random undersampling to mitigate class imbalance.

Performances were evaluated on the outer test folds using F1 score, sensitivity, specificity, accuracy, balanced accuracy, and the area under the ROC curve (AUC). ROC curves were averaged across folds to report mean AUC with standard deviation. The models' ROC curves were compared against the ESC score \cite{OMahony2014} computed on the same data partitions.

For external validation, the five outer-fold models of each classifier were combined into ensemble by calculating the mean of their probabilities and applied to the Rennes cohort.
The best models in terms of AUC on the nested CV were selected for detail analysis and external validation. 
Figure  \ref{fig:methodo_long_analysis} illustrates the CV evaluation (top part) as well as the procedure to create the ensemble model (bottom part). 

\subsection{Feature importance analysis \& explainable AI }
Explainability was a central element of this study. SHAP (SHapley Additive exPlanations) analysis was used to quantify the contribution of each feature to individual predictions \cite{Lundberg2017}. 
 
For the following classifiers: random forest, support vector machines, and gradient boosting, SHAP values were computed on each submodel of the ensemble model.
To derive a global representation of feature importance, the SHAP values and coefficients from the five cross-validation folds were pooled and averaged, ensuring the final rankings reflected the model's behavior.
For the logistic regression, the coefficients are directly accessible for each submodel of the ensemble model.


\subsection{Longitudinal Risk Follow-up}
To explore temporal patterns in predicted risk, a longitudinal analysis was conducted in patients with repeated echocardiographic exams. For each train and test outer fold, the model was applied to all earlier exams available for each patient in the corresponding test set, generating a probability of experiencing the composite outcome at 5 years for each time-point. This enabled the reconstruction of individual risk trajectories. For all the patients with at least two exams, we computed the slope over the predicted values to determine whether the model captured increasing risk prior to events. The absolute value of the slope were also analyzed to be able to depict patients with or without any evolution.
\vspace{1cm}

\section{Results}

\subsection{Model Training}

\begin{figure}[ht]
\centering
\includegraphics[width=8.8cm]{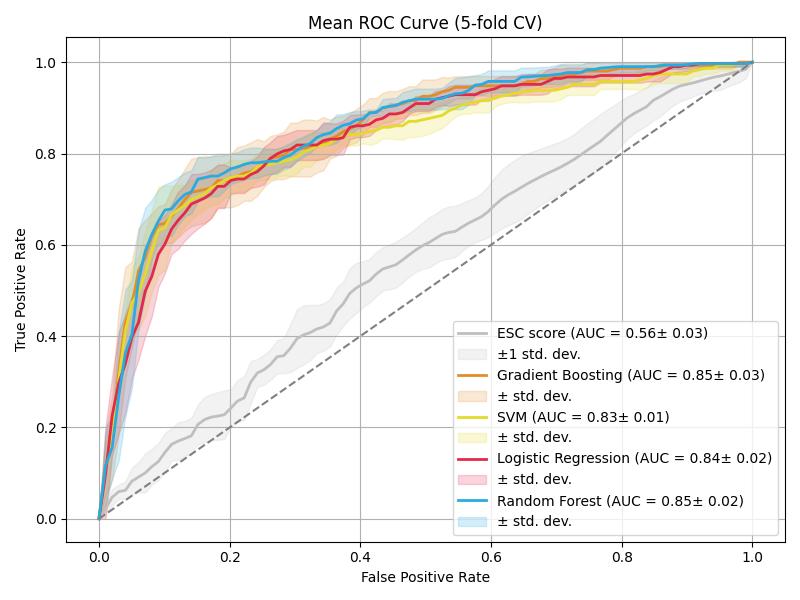}
\caption{Mean ROC curves of the nested cross-validation of the LR, GB, RF and SVM models compared against the ESC risk score.}
\label{fig:CV_all_models}
\end{figure}

Following the initial feature selection phase (correlation analysis), 33 features were retained for subsequent model training.
Figure \ref{fig:CV_all_models} shows the ROC curves for the four ML algorithms compared against the ESC score. RF presents the best AUC with $0.85 \pm 0.03$. It is followed by GB ($0.85 \pm 0.03$) and LR ($0.84 \pm 0.02$), and then SVM ($0.83 \pm 0.02$), with the ESC score arriving last at $0.56 \pm 0.03$.
Table \ref{tab:CV_metrics} regroups the mean and standard deviation of the following metrics: sensitivity, specificity, balanced accuracy, and F1-score of the four models on the nested CV. RF reached the highest F1 score ($0.656 \pm 0.027$), followed closely by GB ($0.644 \pm 0.036$) and SVM ($0.645 \pm 0.028$). A Friedman test was conducted across the model outputs, revealing that the differences in predicted probabilities between the architectures are statistically significant ($p < 0.05$)
RF was chosen for external validation and explainability due to its F1 score as well as its transparency and simplicity, which are beneficial for clinical interpretability.

\subsection{Feature importance analysis}

\begin{figure}[ht]
\centering
\includegraphics[width=8.7cm]{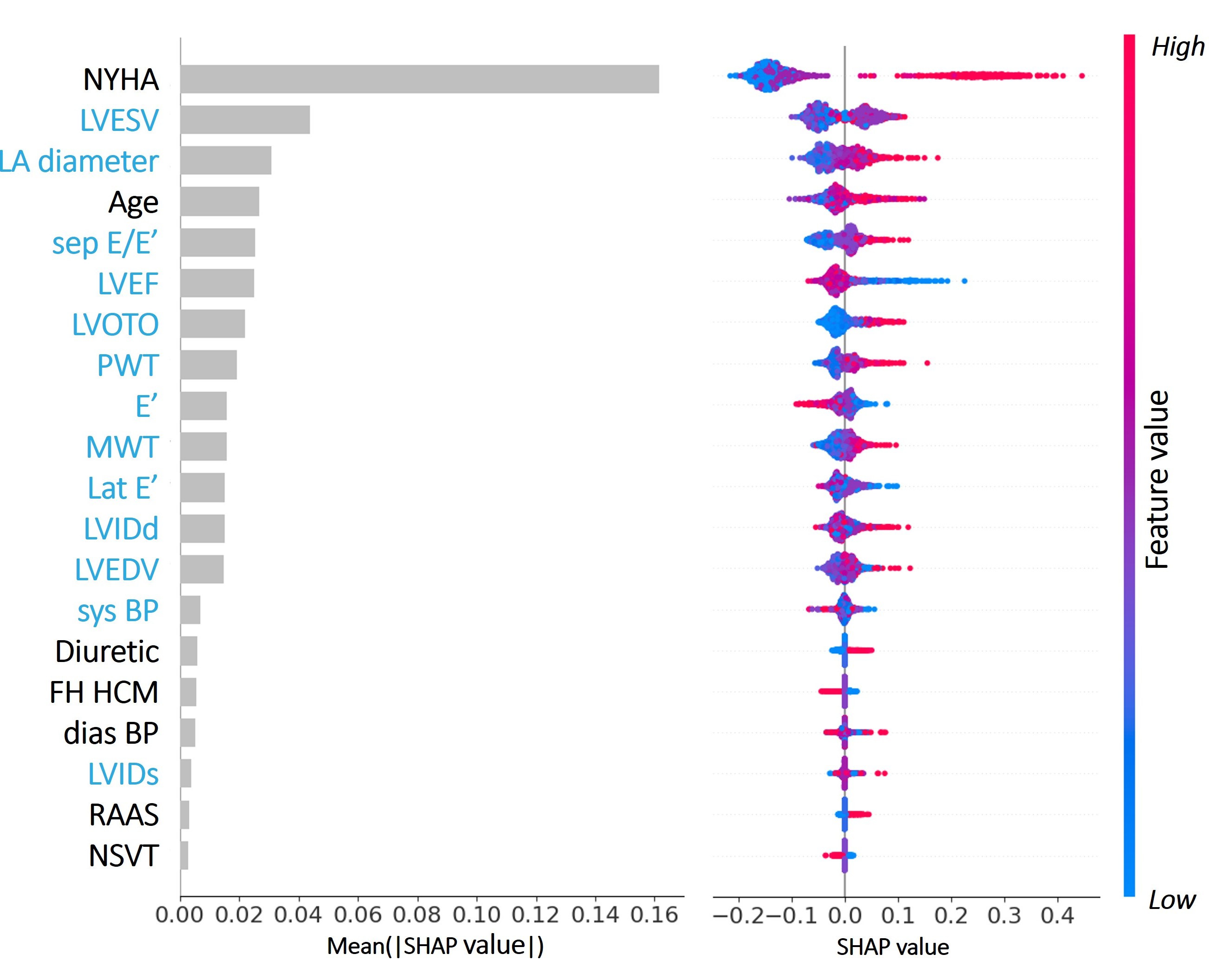}
\caption{Average SHAP summary plot of the global feature importance and directional impact for the five folds of the RF model. Echocardiographic features are written in blue. NYHA: New-York Heart Association index, LVESV or LVESDV: Left Ventricle End Systolic/Diastolic Volume, LVEF: Left Ventricle Ejection Fraction, PWT: Posterior Wall Thickness, sep E/E': ratio of the E wave with septal E', E':peak E' velocity, LA diameter: Left Atrium diameter, LVOT, LV Outflow Tract Obstruction, LVIDd/s: LV  Internal Dimension at end-diastole/systole, MWT: Maximum Wall Thickness, lat E': lateral E', dias/sys BP: diastolic/systolic Blood Pressure, FH HCM: family history of HCM, NSVT: non sustained ventricular tachycardia, RAAS : Renin-angiotensin-aldosterone system medication.}
\label{fig:Feat_importance}
\end{figure} 

Figure \ref{fig:Feat_importance} shows the average feature importance over the five folds of the nested CV of the RF thank to SHAP analysis. The features are ordered by importance, and the color and location around the vertical line represent their impact (\textit{right panel}). For example, the NYHA functional class demonstrates a clear positive correlation with predicted risk: high feature values (indicated in pink) are predominantly located to the right of the vertical baseline, signifying that a higher NYHA class increases the probability of an event. Conversely, low and median feature values (indicated in blue and purple) are clustered to the left of the baseline, representing a negative impact on the risk score. The most important features are consistent with established risk factors, including the presence in high positions of Left Ventricular Ejection Fraction (LVEF), which is a key component of the AHA guidelines \cite{Ommen2020}, and Left Atrial (LA) diameter, Left Ventricular Thickness (LVT), and age, all factors included in the ESC risk score \cite{OMahony2014}. 
Additionally, the analysis highlights the relevance of clinical measurements such as cuff blood pressure. Other echo-extracted features not explicitly included in current guidelines, specifically those involving the heart valves (e.g., mitral valve abnormalities such as regurgitation and stenosis) and diastolic function (e.g., markers like septal E/E' and Left Ventricular End-Diastolic Volume (LVEDV), are also presented, along with posterior wall thickness (PWT). In a lower importance level, variables related to medication or patient history are also represented.

The separated figures for the 5 other folds are provided in supplementary material. Feature importance patterns were consistent across folds.

\begin{table*}[h!]
    \centering
    \caption{Sensitivity, specificity,  accuracy F1 score and AUC of the 4 types of model on nested-CV (SHARE dataset): Gradient Boosting (GB), Support Vector Machine (SVM), Logistic Regression (LR) and Random Forest (RF). The last row correspond to the ensemble RF on the external validation cohort (Rennes).}
    \label{tab:CV_metrics}
    \begin{tabular}{l|ccccc}
Model & Sensitivity & Specificity & Balanced Accuracy & F1-Score & AUC\\
\hline
        RF & $0.738 \pm 0.042$ & $0.823 \pm 0.033$ & $0.780 \pm 0.019$ & $0.656 \pm 0.027$ & $0.852 \pm 0.017$ \\
        GB & $0.725 \pm 0.060$ & $0.818 \pm 0.012$ & $0.771 \pm 0.029$ & $0.644 \pm 0.036$ & $0.850 \pm 0.027$ \\
        LR & $0.767 \pm 0.056$ & $0.763 \pm 0.030$ & $0.765 \pm 0.015$ & $0.626 \pm 0.014$ & $0.837 \pm 0.020$ \\
        SVM & $0.731 \pm 0.043$ & $0.813 \pm 0.040$ & $0.772 \pm 0.020$ & $0.645 \pm 0.028$ & $0.830 \pm 0.015$ \\\hline
        Ensemble RF & 0.351 & 0.877 & 0.614 & 0.342 & 0.723\\ 
        \hline

    \end{tabular}
\end{table*}

\subsection{Longitudinal analysis}

\begin{figure*}[ht]
\centering

\includegraphics[width=18cm]{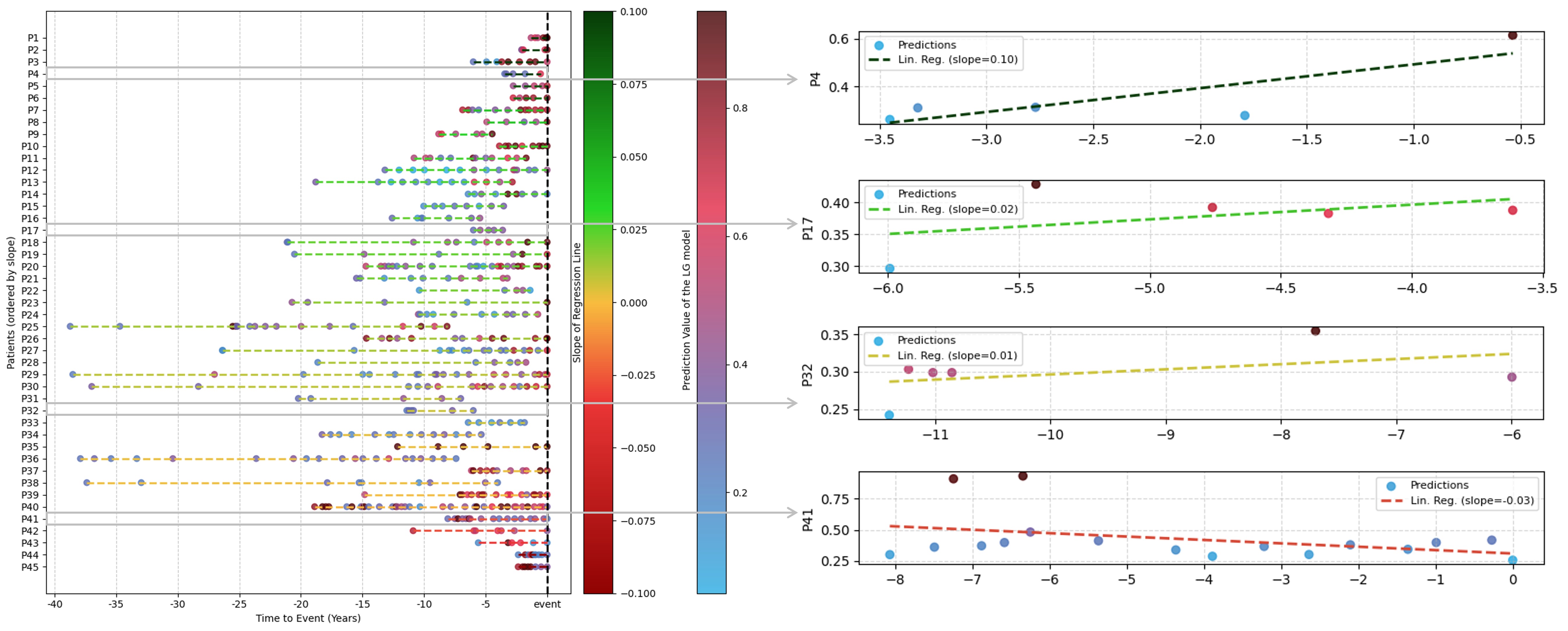}

\caption{Longitudinal analysis visualization: the model's prediction value (blue-to-red scatters) at each exam and the linear regression colored by their slope value (red-to-green).Only patients of the first test set fold who experienced endpoint with at least 5 exams are plotted (P1-P45). Four patients (P4, P17, P32 and P41) were zoomed to show the dynamic of the slopes.}
\label{fig:long_ana}
\end{figure*}

To explore temporal patterns in predicted risk, we conducted a longitudinal analysis in patients with repeated echocardiographic assessments. The model trained on the first outer fold was applied to all earlier exams available for each patient in the corresponding test set, generating a probability of experiencing the composite outcome at 5 years for each timepoint.
Figure \ref{fig:long_ana} visualizes these trajectories for patients who experienced the composite endpoint. While the quantitative analysis included all patients with at least two longitudinal exams, the figure displays only those with five or more assessments to ensure visual clarity. In this plot, individual model predictions are represented by dots following a blue-pink-red color scale, where the shift toward red indicates an increasing predicted probability over time. The overall trend for each patient is summarized by a linear regression slope, visualized with a red-orange-green scale to differentiate the magnitude and direction of risk evolution.

Results revealed a distinct divergence in risk evolution; specifically, 67\% of patients who experienced the endpoint exhibited an upward risk trajectory prior to the event (mean slope $0.1448 \pm 0.7512$ p/year). In contrast, patients who remained event-free showed a significantly more stable profile, with a mean slope of $0.0032 \pm 0.0855$ p/year. These findings suggest that the ML model can effectively capture progressive clinical deterioration over time, demonstrating its potential utility as a dynamic monitoring tool for personalized patient follow-up.

The same analysis was performed on the remaining four cross-validation folds. To provide a comprehensive assessment, the results were aggregated by pooling the calculated slopes from all five folds into a single distribution. Combining the results across all five cross-validation folds demonstrates a measurable difference in the dynamic risk profile. Specifically, patients who experienced the composite outcome showed a significantly higher mean positive slope for their predicted risk score ($0.0431 \pm 0.4578$ probability/year). In contrast, event-free patients exhibited a predicted risk evolution with a slope close to zero ($-0.0027 \pm 0.0664$ probability/year). This difference in the mean slope between the two groups was statistically significant ($T\text{-test } p=0.02745$). Analyzing the absolute slope values further reinforced this separation, showing that the event population experienced larger changes in risk ($0.1033 \pm 0.4481$) compared to the stable event-free population ($0.0332 \pm 0.0575$), with this difference being highly significant ($T\text{-test } p=0.00057$). This consistency demonstrates that the model may capture progressive deterioration, supporting its utility for patient follow-up and dynamic risk monitoring.

\subsection{External validation}

\begin{figure*}[ht]
\centering
\includegraphics[width=16.cm]{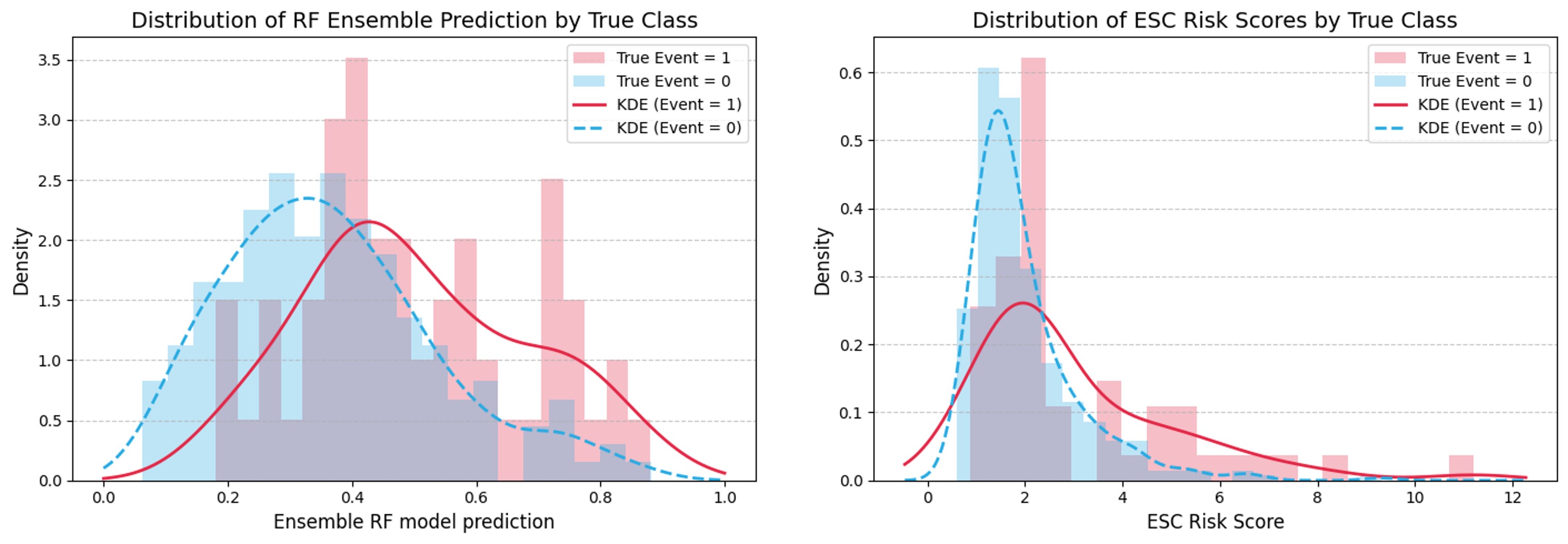}
\caption{Distribution of RF ensemble mean prediction on the external validation population (Rennes) separated depending on the true label (event 0/1), compared with the ESC score (\textit{right}). Kernel Density Estimations (KDE) were plotted over both histograms.}
\label{fig:distrib_pred}
\end{figure*}

\begin{figure*}[ht]
\centering
\includegraphics[width=16.cm]{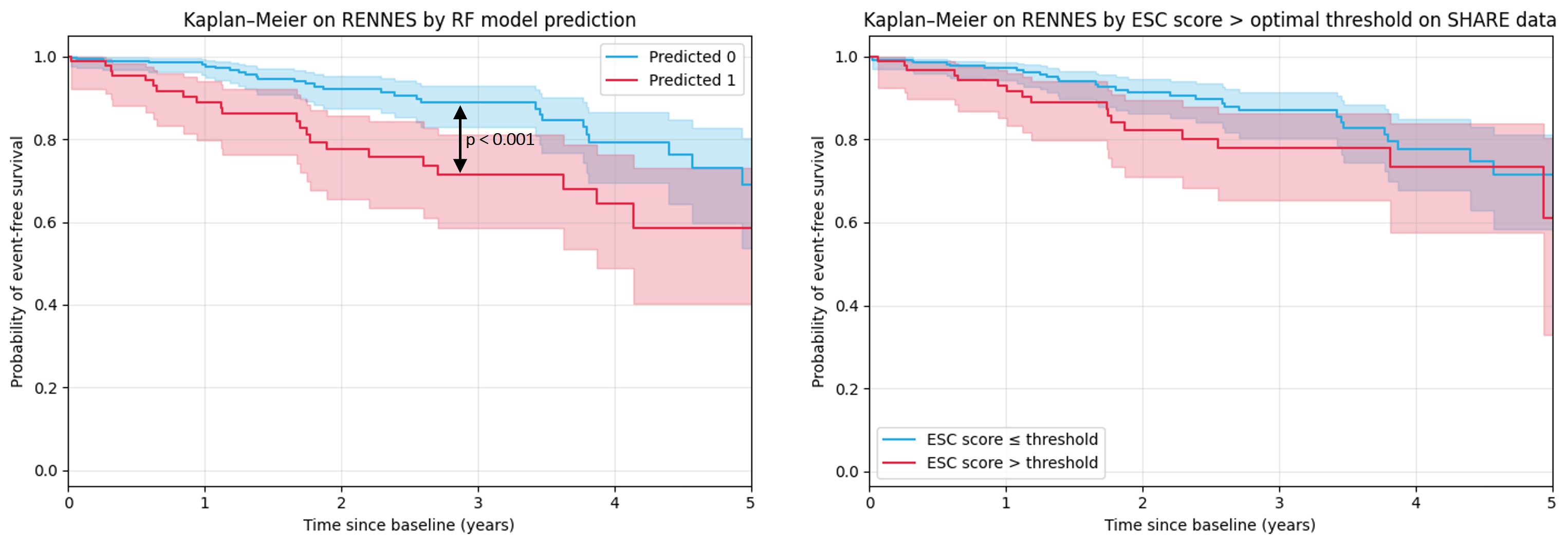}
\caption{Survival curves of the external validation population (Rennes) separated depending on the ensemble RF model prediction (mean probability) at 5 years (predicted 0/1) and ESC score (threshold to separate the population computed on the SHARE dataset).}
\label{fig:survival analysis}
\end{figure*}

After being trained on the Florence database, the five models of each type, created during the nested-CV were grouped in an ensemble model. 
Ensemble RF achieved balanced accuracy of 0.614 on the external Rennes data, with AUC 0.723. While performance decreased compared with internal validation, the models retained clinically meaningful discriminative ability.
Notably, the model maintained a high specificity of 0.806 on the external validation set. This is a significant result as it demonstrates that the model's ability to accurately identify and reject negative cases (low-risk patients) remains robust across different cohorts, avoiding an increase in false-positive risk predictions.
Metrics are shown in the last row of Table \ref{tab:CV_metrics} 
As expected by the similar results on the training phase, the other ensemble models created with LR, GB or SVM have very similar performances. Nevertheless, the performances drop between the nested-CV and the external validation.  

Figure \ref{fig:distrib_pred} illustrates the distribution of the predicted probability scores on the external validation set, comparing patients who experienced a cardiovascular event (True Event~=~1) against event-free patients (True Event~=~0). To generate this continuous distribution plot, the prediction probability was calculated as the mean of the probabilities output by the five individual RF models. For the Ensemble RF model, the distribution of high-risk predictions is significantly shifted toward the event group, confirming the model's discriminative power (e.g., Mann-Whitney U test $p$-value of $2.4 \times 10^{-7}$). In contrast, the distribution of the continuous ESC score is also shown for comparison; while its distribution is also shifted, the separation between the two patient groups is less pronounced (e.g., Mann-Whitney U test $p$-value of $1.7 \times 10^{-5}$).

\subsection{Survival analysis \& comparison with the ESC score}

Figure \ref{fig:survival analysis} presents the survival curves for the external validation population, separated according to the RF ensemble model prediction and the ESC score threshold. Despite the moderate predictive performance of the RF ensemble model on the external validation set using classical metrics, the classification it provides demonstrates significant value in risk stratification. The survival curve analysis based on the RF ensemble prediction shows a clear separation between the two populations (patients predicted by the model to be risk '0' or risk '1'), supported by a highly significant Log-rank test p-value of $8.62 \times 10^{-4}$. In contrast, the ESC score produced a weaker separation of the populations ($p = 0.0559$), confirming that the machine learning-based classification is superior at identifying high-risk patients. Because the ESC score provides a continuous risk probability rather than a binary classification, we chose the optimal threshold (the point closer to the upper-left corner) determined from the ROC curve computed on the training set (Figure \ref{fig:CV_all_models}) to separate the external validation population into high- and low-risk groups.

\section{Discussion} 

This study introduces a clinically adapted, explainable machine learning (ML) score for 5-year cardiovascular risk prediction in Hypertrophic Cardiomyopathy (HCM) that integrates routinely available echocardiographic, clinical, and medication data. The model demonstrated strong internal discriminative performance (AUC $0.85 \pm 0.02$) and, crucially, maintained a high capacity for risk stratification in an independent external validation cohort, significantly outperforming the conventional ESC score in survival curve separation ($\text{Log-rank } p<0.001$).
This finding is consistent with contemporary research demonstrating that sophisticated multimodal AI approaches, such as \cite{Lai2025, Zhao2024} on cardiac magnetic resonance (CMR) and electronic health records (EHR), achieve significantly higher AUCs than current clinical guidelines.While these studies highlight the power of integrating tissue-characterization information from CMR, our approach focuses on widely available echocardiographic parameters to facilitate clinical integration into routine workflows where advanced imaging may be less accessible (cfr. Table \ref{tab:prior_ml_approaches}).

Our use of multimodal AI-driven approaches to improve risk stratification is consistent with current research directions in HCM, which emphasize the analysis of complex data from echocardiography, ECG, and clinical assessments \cite{Panichella2025a}.
The model’s simplicity is its main strength; all required variables are routinely collected and accessible in EHR, facilitating its adoption. The high interpretability provided by the explanation framework supports clinical trust, aids in shared decision-making, and facilitates patient communication.

This work represents the first study to investigate the longitudinal evolution of a predicted ML-based risk score in HCM. Our analysis demonstrated a measurable increase in the predicted risk score slope for patients who subsequently experienced an event, while event-free patients exhibited stable scores with slopes close to zero. This temporal dynamic supports the utility of the ML score not only for baseline prediction but also for dynamic risk monitoring during routine follow-up. While other studies have explored ML for adverse-event prediction in HCM, reporting improved discrimination over traditional risk factors, these studies were often limited by small cohorts and the absence of external validation \cite{Kochav2021}. Our study addresses this critical gap by successfully validating the model on an independent, external cohort and providing a framework for longitudinal monitoring, an area that has seen limited investigation in prior ML HCM research.

The risk prediction is highly dependent on a multitude of factors, as demonstrated by the feature importance analysis. The evolution of the predicted risk score over time reflects changes in these underlying clinical features. It must be emphasized that these clinical features, such as left ventricular outflow tract gradient or functional capacity (NYHA class), can be significantly influenced by medical intervention, particularly the use of $\beta$-blockers or calcium channel blockers \cite{Nistri2012, Borlaug2022}. Therefore, the change in a patient's predicted risk score over time depends not only on underlying disease progression but also on the response to and change in medication, highlighting the complex, non-static nature of risk in this population.

Our analysis of the contribution of basic ECG extracted features (heart rate and interval durations: QT, PR, QRS) showed limited incremental value. While previous conference work \cite{taconne2025ESC} suggested an added predictive value for these simple markers, this was not reflected in the final performance on external validation. Consequently, the results detailing the contribution of these basic ECG features are not reported here. This suggests that more complex ECG features related to arrhythmic substrate or LVH pathology \cite{Taconne2025b, Lyon2018} could yield greater increases in risk prediction accuracy. However, extracting these often necessitates access to the raw ECG signal, which is less common in readily available structured EHR data compared to simple interval durations.

Regarding patient demographics, while age is an included feature, it does not consistently appear as the primary determinant of risk in our final model, especially when compared to key echocardiographic and clinical features. This observation is consistent with the understanding that while risk increases with age in HCM, the presence of specific structural and symptomatic markers are often more immediate and actionable predictors of adverse outcomes. We also noted a gender imbalance between the Florence (40\% female) and Rennes (34\% female) cohorts; however, testing showed that the exclusion of gender information did not lead to significant changes in model performance. This lack of influence is further supported by the SHAP analysis, where sex consistently appears as one of the least important features utilized by the algorithm.

Our machine learning approach also provided valuable, hypothesis-free insights into the underlying determinants of adverse outcomes. Top predictors, consistently highlighted by the SHAP analysis, demonstrated strong coherence with established clinical criteria: Features such as LV diameter, maximum wall thickness, LVOT, and age ranked highly, aligning directly with key variables used in the European Society of Cardiology (ESC) risk score. Consistent with American Heart Association (AHA) guideline criteria for risk assessment, LVEF (Left Ventricular Ejection Fraction) also ranked as a top predictor. The analysis further identified several non-guideline parameters as important, including: mitral and tricuspid valve abnormalities (regurgitation and stenosis), markers of diastolic dysfunction (septal E', LVEDV), and posterior wall thickness. These results suggest the model effectively integrates both established risk factors and novel echocardiographic markers with prognostic relevance, offering a broader characterization of patient risk profiles

The primary limitations of this study include the exclusive use of European hospital-based cohorts, which may limit generalizability to other populations, and the use of a composite endpoint that combines arrhythmic and heart failure events. Additionally, while the model proved robust for relative risk stratification, the external validation performance was likely affected by disparities in database construction and the shorter follow-up period in the Rennes cohort (2.3 vs 5 years). These results suggest that while the general risk structure is preserved across centers, recalibration may be required to optimize absolute risk estimation in new populations.

Future work should explore recalibration methods to improve external validation performance. Evaluating the incremental value of the ML score over existing scores, will require a larger, multicenter database to ensure sufficient statistical power and robustness. 
Furthermore, efforts should be directed toward incorporating more complex features directly linked to LVH severity or specific to HCM pathology, extracted from echocardiography such as strain \cite{Galli2019, Wazzan2023, AlWazzan2024} or CMR markers of fibrosis \cite{Fahmy2022, Kiaos2024, Zhao2024}.
Prospective validation is warranted to evaluate the clinical utility and the real-world benefit.

\section{Conclusion} A simple, explainable machine learning score, leveraging routinely acquired multimodal data, provides a meaningful and interpretable approach to risk stratification for HCM patients. The model demonstrates superior separation of high- and low-risk groups compared with the ESC score on external validation, and offers a novel capacity for dynamic risk monitoring throughout patient follow-up. These features highlight the potential of data-driven methods to enhance precision in clinical decision-making for HCM management. 

\appendices

\section*{Acknowledgment}

This research study is part of the project INSIGHT-LV and SMASH-HCM funded by the European Union under the Marie Skłodowska-Curie GA \#101198472 and GA \#101137115 respectively.

\section*{References}

\bibliographystyle{ieeetr}

\bibliography{library.bib}

\end{document}